\documentclass[11pt]{article}
\usepackage{coling2020}
\usepackage{times}
\usepackage{url}
\usepackage{latexsym}

\usepackage{amsmath}
\usepackage{amsfonts}
\usepackage{mathrsfs}
\usepackage{amssymb}
\usepackage{url}
\usepackage{pgfplots}
\usepackage{subfigure}
\usepackage{multirow}
\usepackage{latexsym}
\usepackage{booktabs}
\usepackage{appendix}
\colingfinalcopy % Uncomment this line for the final submission

\title{Revisiting Simple Domain Adaptation Methods \\ in Unsupervised Neural Machine Translation}

\author{ Haipeng Sun{$^1$}\thanks{\;\;Haipeng Sun was an internship research fellow at NICT when conducting this work.}, Rui Wang{$^2$},  Kehai Chen{$^2$}, Masao Utiyama{$^2$},\\ \textbf{Eiichiro Sumita{$^2$}, Tiejun Zhao{$^1$}, and Chenhui Chu{$^3$}} \\
	$^1$Harbin Institute of Technology, Harbin, China \\
	$^2$National Institute of Information and Communications Technology (NICT), Kyoto, Japan \\	$^3$Osaka University, Osaka, Japan \\
	\texttt{hpsun@hit-mtlab.net}, \texttt{tjzhao@hit.edu.cn}, \texttt{chu@ids.osaka-u.ac.jp} \\
	\texttt{\{wangrui, khchen, mutiyama, eiichiro.sumita\}@nict.go.jp} \\}

\date{}

\begin{document}
\maketitle
\begin{abstract}
 %Neural machine translation (NMT) has been prominent in many machine translation tasks. However, in some domain-specific tasks, only the corpora from similar domains can improve translation performance. If out-of-domain corpora are directly added into the in-domain corpus, the translation performance may even degrade. Therefore, domain adaptation techniques are essential to solve the NMT domain problem.
 Domain adaptation has been well-studied in supervised  neural machine translation (SNMT). 
 However, it has not been well-studied for unsupervised neural machine translation (UNMT), although UNMT has recently achieved remarkable results in several domain-specific language pairs.
 %In  a  real-world  scenario,  it is  difficult  to  get  enough  in-domain parallel or even monolingual corpora in some other specific domain.  
 Besides the domain inconsistence between parallel training data and test data for SNMT, there sometimes exists domain inconsistence between two monolingual training data for UNMT. 
 In this work, we empirically categorize different domain adaptation scenarios for  UNMT. 
 Based on these scenarios, we revisit the effect of the existing representative domain adaptation methods including batch weighting and fine tuning methods in UNMT. Finally, we propose modified methods to improve the performances of domain-specific UNMT systems.
\end{abstract}

\section{Introduction}
Neural Machine Translations (NMT) have set several state-of-the-art new benchmarks \cite{bojar-etal-2018-findings,barrault-etal-2019-findings}. 
Recently, unsupervised NMT (UNMT) has attracted great interests in the machine translation community~\cite{DBLP:journals/corr/abs-1710-11041,lample2017unsupervised,P18-1005,lample2018phrase,sun-etal-2019-unsupervised}.
Typically, UNMT relies solely on monolingual corpora in similar domain rather than bilingual parallel data for supervised NMT (SNMT) to model translations between the source language and target language and has achieved remarkable results on several translation tasks~\cite{DBLP:journals/corr/abs-1901-07291}.

The available training data is ever increasing; however, only the related-domain corpora, also called in-domain corpora, are able to improve the NMT performance \cite{koehn-knowles-2017-six}. Additional unrelated corpora, also called out-of-domain
corpora, are unable to improve or even harm the NMT
performance for some domains such as TED talks and some
tasks such as IWSLT \cite{wang-etal-2017-instance}.%should explain what the in-domain MT task is

%s between source language and target language, and has achieved remarkable results on several translation tasks~\cite{DBLP:journals/corr/abs-1901-07291}.

%However, two monolingual training data for UNMT are often different domains in the real-world scenario, which makes UNMT face a congenital defect of domain adaptation.
Domain adaptation methods have been well-studied in SNMT \cite{chu-etal-2017-empirical,DBLP:conf/aclnmt/ChenCFL17,wang-etal-2017-sentence,wang-etal-2017-instance,DBLP:conf/emnlp/WeesBM17,DBLP:conf/wmt/FarajianTNF17,DBLP:conf/coling/ChuW18} while they have not been  well-studied in UNMT.
%{\color{red}In contrast, domain adaptation methods for UNMT have never been proposed.是否需要}
For UNMT, in addition to inconsistent domains between training data and test data for SNMT, there also exist other inconsistent domains between monolingual training data in two languages.
Actually, it is difficult for some language pairs to obtain enough source and target monolingual corpora from the same domain in the real-world scenario.
%Compared with the inconsistent domain between training data and test data for NMT, there exists an inconsistent domain between two monolingual training data for UNMT.
In this paper, we first define and analyze several scenarios for 
%%unsupervised neural machine translation 
UNMT with specific domain.
%\begin{itemize}
%\item  1: Resource-rich out-of-domain monolingual corpora and resource-poor in-domain monolingual corpora for both languages.
%\item  2: No  in-domain monolingual corpora for one language.
%\item  3: No resource-rich out-of-domain monolingual corpora for one language.
%\item  4: Resource-rich out-of-domain monolingual corpora for one language and resource-poor in-domain monolingual corpora for one language.
%\end{itemize}
%The other scenarios such as no resource-rich out-of-domain or no in-domain monolingual corpora for both languages would be not taken into consideration in this paper. 
On the basis of the characteristics of these scenarios, we revisit the existing domain adaptation methods including batch weighting  and fine tuning methods in UNMT. Finally, we proposed modified domain adaptation methods  to improve the performance of UNMT in these scenarios. 

%again, better to be specific and briefly introduce the results of existing methods and modified methods here
To the best of our knowledge, this paper is the first work to explore  domain adaptation problem in UNMT.
%This paper primarily makes the following contributions:

%\begin{itemize}
%\item  We empirically introduce various scenarios for domain-specific UNMT. Some scenarios are unique scenarios for UNMT domain adaptation.
%\item We propose several potential solutions, achieving up to 17 BLEU scores improvement for UNMT.
%\end{itemize}

\begin{table*}[th]
	
	\centering
	\scalebox{.88}{
		\begin{tabular}{c|l|cccc}
			\toprule
			Scenarios & Abbreviation & $L_1$ in-domain & $L_2$ in-domain & $L_1$ out-of-domain & $L_2$ out-of-domain \\
			\midrule
			\multirow{3}{*}{\begin{tabular}[c]{@{}l@{}}Monolingual corpora  \\from same domains\end{tabular}} & $II$ &\checkmark &\checkmark &$\times$& $\times$\\
			\cline{2-6}
			& $OO$ & $\times$& $\times$&\checkmark &\checkmark\\
			\cline{2-6}
			& $IIOO$ &\checkmark &\checkmark &\checkmark &\checkmark\\
			\midrule
			\multirow{6}{*}{\begin{tabular}[c]{@{}l@{}}Monolingual corpora \\ from different domains\end{tabular}} & \multirow{2}{*}{$IOO$}  &$\times$ & \checkmark &\checkmark& \checkmark\\
			\cline{3-6}
			 & &\checkmark & $\times$ &\checkmark& \checkmark\\
			\cline{2-6}
			& \multirow{2}{*}{$IIO$} &\checkmark &\checkmark & \checkmark& $\times$\\
			\cline{3-6}
			&  &\checkmark &\checkmark & $\times$& \checkmark\\
			\cline{2-6}
			& \multirow{2}{*}{$IO$} & $\times$ &  \checkmark &\checkmark& $\times$\\
			\cline{3-6}
			&  & \checkmark&  $\times$  &$\times$ & \checkmark\\
			\bottomrule
	\end{tabular}}
	\caption{The statistics of monolingual training corpora for different scenarios. $\checkmark$ denotes having this monolingual corpus in one scenario; $\times$ denotes having no this monolingual corpus in one scenario.}
	\label{tab:scenario}
\end{table*}
\section{UNMT Domain Adaptation Scenarios}
\label{scenarios}
%Compared with domain adaptation for NMT relies on parallel corpora, 
In SNMT, all the corpora are parallel and the domains of source and target corpora are the same. Therefore, the domain adaptation technologies focus on the domain shift between the training and test corpora. In UNMT, there is only monolingual corpora and the domains of source and target corpora are sometimes different. Therefore, there are more scenarios of UNMT domain adaptation.
Given two different languages $L_1$ and $L_2$, we define two main scenarios according to the domains of two languages in the training set: monolingual training corpora from the same domain, and monolingual training corpora from different domains, as shown in Table \ref{tab:scenario}.

Take monolingual corpora from different domains as an example, we further divide this scenario into three sub-scenarios: $IOO$, $IIO$, and $IO$, where ``$I$" denotes the in-domain data for one language and ``$O$" denotes the out-of-domain data for one language. Further, $IOO$ denotes there are resource-rich out-of-domain monolingual corpora for both languages and resource-poor in-domain monolingual corpora for language $L_2$. Especially, we regard ``$L_2$ in-domain + $L_1$ out-of-domain" and ``$L_1$ in-domain + $L_2$ out-of-domain" as the same scenario $IO$. Note that scenario $II$ and $OO$ were only as the baselines to evaluate other four scenarios. In this paper, we consider other four scenarios to improve translation performance.

%According to our introduced scenarios, we propose three simple methods, that is, batch weighting, data selection, and fine tuning. Figure \ref{fig:UNMT} describes the training relationship and procedure for all scenarios.  Specifically, data selection method usually works with fine tuning method to improve translation performance.

\section{Domain Adaptation Methods}
According to our introduced scenarios, we revisited two simple  domain adaptation methods, that is, batch weighting and fine tuning.
%better to explain the reason why you chose these two among many

%Table \ref{tab:method} describes the training relationship and procedure for all scenarios.  %Specifically, data selection method usually works with fine tuning method to improve translation performance.

\subsection{Batch Weighting for UNMT}
\label{Weighting}

%the difference in the number of two monolingual corpora would cause over-fitting in one direction when it does not complete training on the other direction. 
%The larger out-of-domain monolingual corpora is not fully utilized.

%\noindent \textbf{Batch Weighting for UNMT}
\textbf{Original:} The batch weighting method \cite{wang-etal-2017-instance} for SNMT  is difficult to be directly transferred to the UNMT training because the training data of the source and target languages are sometimes unbalanced (such as the $IO$ and $IIO$ scenarios). %for readers to understand this, a brief description of batching weighting maybe with equations for SNMT is required
Regardless of training cross-lingual language model or UNMT model, the model causes over-fitting in one language which includes the smaller amount of in-domain monolingual corpus.
In other words, the large-scale out-of-domain monolingual corpus for other language is not fully utilized. 

\noindent \textbf{Modified:} %this naming sounds a trivial difference, better to give a specific name 
To address this issue, we propose a batch weighting method for UNMT domain adaptation to make full use of out-of-domain corpus to build a robust UNMT model when there exists only one large-scale out-of-domain monolingual corpus in one scenario. 
Specifically, we adjust the weight of out-of-domain sentences to increase the amount of out-of-domain sentences rather than to increase that of in-domain sentences~\cite{8360031} in every training batch.
%In contrast to the method of \newcite{wang-etal-2017-instance}, we adjust the weight of out-of-domain sentences to increase the amount of out-of-domain sentences in every training batch 
%so that the larger out-of-domain monolingual corpora can be fully utilized. 
In our batch weighting method, the out-of-domain sentence ratio is estimated as 
\begin{equation}
	\begin{aligned}
		\mathcal{R}_{out} &=\frac{\mathcal{N}_{out}}{\mathcal{N}_{out}+\mathcal{N}_{in}},
		%\frac{|\mathcal{M}_{out}|}{|\mathcal{M}_{in}|+|\mathcal{M}_{out}|}
		%&=,
		\label{eq:rout}
	\end{aligned}
\end{equation}
%where $|\mathcal{M}_{out}|$ and $|\mathcal{M}_{in}|$ are the sentence number loaded from out-of-domain and in-domain monolingual corpora in every batch, respectively. 
where $\mathcal{N}_{in}$ is the number of mini-batches loaded from in-domain monolingual corpora in intervals of $\mathcal{N}_{out}$ mini-batches loaded from out-of-domain monolingual corpora.
%For example, .
%Compared with every batch of normal UNMT model training including two mini-batch loaded from different language monolingual corpora, every batch of UNMT domain adaptation model containing $1$ mini-batch loaded from in-domain monolingual corpus and $\mathcal{N}$ mini-batch loaded from out-of-domain monolingual corpus after introducing batch weighting method.

For the $IO$ and $IIO$ scenario, we apply the proposed batch weighting method to train cross-lingual language model and UNMT model in turn since the quantity of training data in two languages is quite different in the $IO$ and $IIO$ scenario. %a briefly introduction of UNMT training is also necessary for readers who are unfamiliar with UNMT
For $IOO$ and $IIOO$ scenario, there are two large-scale out-of-domain monolingual corpora and their quantity is similar. Therefore, batch weighting method is not so necessary for these scenarios.
\subsection{Fine Tuning}
\label{FineTuning}
\textbf{Original:}
%We apply the existing fine tuning method to UNMT models under the $IIOO$, $IIO$ scenarios, thus exploring the solution of domain adaptation. 
For the $IIOO$ and $IIO$ scenarios, we first train  UNMT model on the corresponding corpora  until convergence. Then we further fine tune parameters of the UNMT model on the resource-poor in-domain monolingual corpora for both languages. However, The original fine tuning method is difficult to directly transferred to the UNMT training under the $IOO$, $IO$ scenarios since there only exist in-domain data for language $L_2$ under these scenarios as shown in Table \ref{tab:scenario}.

\noindent \textbf{Modified:} We propose modified data selection method \cite{moore-lewis-2010-intelligent,axelrod-etal-2011-domain} to select pseudo in-domain data from out-of-domain data for another language $L_1$.
The traditional data selection for SNMT domain adaptation \cite{wang-etal-2017-sentence,8360031} is not suitable for UNMT because in-domain language model could not be trained where does not exist in-domain corpus for language $L_1$.

To address this issue, we back-translate the language $L_2$ in-domain data to language $L_1$ pseudo in-domain data, using an UNMT baseline system. Then, we use these corpora to train a cross-lingual language model as the in-domain language model. For the $IO$ scenario that just exists out-of-domain corpus for  language $L_1$ as shown in Table \ref{tab:scenario}, we randomly select language $L_1$ out-of-domain corpus that is similar in size to the language $L_2$ in-domain corpus and take the same approach to train a cross-lingual language model as the out-of-domain language model. For the $IOO$ scenario that exist out-of-domain corpora for both languages, we randomly select out-of-domain corpora that are similar in size to the language $L_2$ in-domain corpus, respectively. Then we train a cross-lingual out-of-domain language model, using these corpora.

In practice, we adopt the data selection method \cite{moore-lewis-2010-intelligent,axelrod-etal-2011-domain}, and rank an out-of-domain sentence $s$ using:
\begin{equation}
	\begin{aligned}
		CE_I(s)-CE_O(s),
	\end{aligned}
\end{equation}
where $CE_I(s)$ denotes the cross-entropy of the sentence $s$ computed by the in-domain language model; $CE_O(s)$ denotes cross-entropy of the sentence $s$ computed by the out-of-domain language model. %what is CED? you have to use this term around Eq (2) beforehand. also better to describe both CED and CE methods there beforehand
This measure biases towards sentences that are both like the in-domain corpus and unlike the out-of-domain corpus. 
Then we select the lowest-scoring sentences as the pseudo in-domain corpus.% to further enhance translation performance. 

%Compared with the previous data selection for NMT domain adaptation, we just select the pseudo in-domain corpus for the language for which there is no in-domain corpus, not for both languages.
Finally, we further fine tune parameters of the UNMT model on the resource-poor in-domain monolingual corpora for language $L_2$ and the pseudo in-domain corpus for language $L_1$ after we apply modified data selection method to achieve the pseudo in-domain corpus for language $L_1$.

%For $IIO$, we further fine tune parameters of the UNMT model on the resource-poor in-domain monolingual corpora for language $L_2$.
%For $IIOO$, we further fine tune parameters of the UNMT on the resource-poor in-domain monolingual corpora for both languages.
%and then fine tune their parameters on the resource-poor in-domain monolingual corpora. 
%The only difference between scenario $IIOO$ and $IOO$ is that there is only $L_2$ in-domain monolingual corpus to fine tune in the scenario $IOO$ compared with two in-domain monolingual corpora in the scenario $IIOO$.

% Finally, UNMT models are gained by applying fine-tuning method to the following two scenarios:
% \begin{itemize}
%     \item UNMT model under the $IIO$: we further fine tune parameters of the UNMT model on the resource-poor in-domain monolingual corpora for language $L_2$.
%     \item UNMT model under the $IO$: we further fine tune parameters of the UNMT on the resource-poor in-domain monolingual corpora for both languages.
% \end{itemize}

%Overall, we proposed batch weighting and fine tuning method for different scenarios as Table \ref{tab:method} shows. These methods would be only used if there are different domain monolingual corpora for both languages. So scenario $II$ and $OO$ would be not taken into consideration in this paper. 
\begin{table}[th]
	\centering
	\scalebox{1}{
		\begin{tabular}{lccc}
			\toprule
			Scenarios & Batch weighting & Fine tuning \\
			\midrule
			$IIOO$ &-  &\checkmark \\
			$IOO$ &- &\checkmark \\
			$IIO$ &\checkmark&\checkmark \\
			$IO$  & \checkmark& \checkmark \\
			\bottomrule
	\end{tabular}}
	\caption{The suitability of the proposed methods for different scenarios. $\checkmark$ denotes that the method is used in this scenario; $-$ denotes that the method is not used in this scenario.}
	\label{tab:method}
\end{table}

Overall, batch weighting method is used in the case that there is no out-of-domain monolingual corpus for one language, including scenario $IIO$ and $IO$; 
%data selection method is used in the case that there is no in-domain monolingual corpus for one language, including scenario $IOO$ and $IO$;
fine tuning method is suitable to all our considered scenarios, including scenario $IIOO$, $IOO$, $IIO$, and $IO$, as shown in Table~\ref{tab:method}.
%Note that scenario $II$ and $OO$ were only as the baselines to evaluate other four scenarios. 
%would be not taken into consideration in this paper.
\section{Experiments}
\subsection{Datasets}
We considered two language pairs to do simulated  experiments on the French (Fr)$\leftrightarrow$English (En) and German (De)$\leftrightarrow$En translation tasks. 
For out-of-domain corpora, we used 50M sentences from WMT monolingual news crawl datasets for each language. 
For in-domain corpora, we used 200k sentences from the IWSLT TED-talk based shuffled training corpora for each language. 
To make our experiments comparable with previous work \cite{8360031}, we reported results on IWSLT test2010 and test2011 for Fr$\leftrightarrow$En and IWSLT test2012 and test2013 for De$\leftrightarrow$En.

For preprocessing, we followed the same method of \newcite{lample2018phrase}. 
That is, we used a shared vocabulary  for both languages with 60k subword tokens based on BPE \cite{sennrich2015neural}. We used the same vocabulary including in-domain and out-of-domain corpora for different scenarios. 
If there exists only one in-domain monolingual corpus in one scenario, we chose Fr/De in-domain monolingual corpus; if there exists only one out-of-domain monolingual corpus in one scenario, we chose En out-of-domain monolingual corpus for uniform comparison.

\subsection{Language Model and UNMT Settings}
We used  the XLM UNMT toolkit\footnote{\url{https://github.com/facebookresearch/XLM}} and followed settings of \newcite{DBLP:journals/corr/abs-1901-07291}. We first trained cross-lingual language model, and followed settings of \newcite{DBLP:journals/corr/abs-1901-07291}: 6 layers for the encoder. The dimension of hidden layers was set to 1024. 
The Adam optimizer \cite{kingma2014adam} was used to optimize the model parameters.
The initial learning rate  was 0.0001,  $\beta_1 = 0.9$, and $\beta_2 = 0.98$. 
We trained a specific cross-lingual language model for each scenario, respectively.
The cross-lingual language model was used to initialize  the encoder and decoder of the whole UNMT model and select pseudo in-domain monolingual corpus.

The UNMT model included 6 layers for the encoder and the decoder. 
The other parameters were the same as that of language model.
We used the case-sensitive 4-gram BLEU score computed by $multi-bleu.perl$ script from Moses \cite{koehn-etal-2007-moses} to evaluate the test sets.
The baselines in different scenarios are the UNMT systems trained on the mixed monolingual corpora including in-domain and out-of-domain data in the corresponding scenarios.

\subsection{Main Results}

\begin{table}
	
	\centering
	\scalebox{0.72}{
		\begin{tabular}{clclcccccccc}
			\toprule
			 \multirow{2}{*}{\#}&\multirow{2}{*}{Scenario} & \multirow{2}{*}{Supervision} &\multirow{2}{*}{Method} & \multicolumn{2}{c}{De-En} & \multicolumn{2}{c}{En-De} & \multicolumn{2}{c}{Fr-En} & \multicolumn{2}{c}{En-Fr}   \\
			&&&&test2012 & test2013 &test2012&test2013&test2010 & test2011 &test2010&test2011\\
			\midrule
			1&\multirow{2}{*}{$II$} &\multirow{2}{*}{Yes}&\newcite{8360031}  &n/a  &n/a&23.07&25.40  &n/a&  n/a&32.11& 35.22\\
			2&&&Base &33.68 &35.41 & 28.09&30.48 &36.13 & 40.07&36.43 &37.58 \\
			\midrule
			3&$II$&\multirow{2}{*}{No}&Base&24.42&25.65&21.99&22.72&25.94&29.73&25.32&27.06\\
			
			4&$OO$&&Base &  21.21&21.66&10.25&9.90&24.28&28.77&23.08&26.08\\
			\midrule
			5&\multirow{2}{*}{$IIOO$} &\multirow{2}{*}{No}&Base&24.87&26.00 &21.64 &22.57 &26.05 &30.18&26.35&30.12\\
			6&&&FT&29.82&31.57&26.48& 28.18 &31.23&35.94&29.08&33.67\\
			\midrule
			7&\multirow{3}{*}{$IOO$} &\multirow{3}{*}{No}&Base& 20.94 & 21.52 & 16.53 & 16.80 & 25.16 &29.88  & 25.18 & 28.73 \\
			8&&&FT(original)&22.75&23.14&21.09&21.78&28.37&33.57&26.16&30.14\\
			%&&DS+FT (original)&&&&&&&&&\\
			%&&DS (modified)&&&&&&&&&\\
			9&&&FT(modified)&24.33  & 24.77 &24.43   & 25.59 & 29.13 & 34.38 & 26.45 &30.69\\
			\midrule
			10&\multirow{3}{*}{$IIO$}&\multirow{3}{*}{No} &Base&11.11 &10.30 &11.54 &11.95 &17.88 &20.32 & 17.02& 18.16 \\
			11&&&FT+BW(original) &19.91&20.19&17.05&17.23 &26.84&29.61&23.18&25.18 \\
			
			12&&&FT+BW(modified)&26.12  & 27.33  &22.63  & 23.72 & 27.88 & 32.16 &25.42  &28.05 \\
			\midrule
			13&\multirow{3}{*}{$IO$} &\multirow{3}{*}{No} &Base& 10.79 & 10.77 & 11.44 & 11.82 &18.00&20.91&16.19&16.84\\
			14&&&BW(original)&8.15&7.05&9.28&9.70&18.00&19.52&16.39&17.72\\
			%15&&&BW(modified)&17.78&18.00&16.01&16.60&22.53&25.29&20.04&22.12\\
			15&&&FT+BW(modified)&19.76  &20.22  & 18.32 & 18.99 & 22.59  & 26.55 & 20.61 & 22.79 \\
			\bottomrule
	\end{tabular}}
	\caption{The BLEU scores in the different scenarios for En-De and En-Fr language pairs. Base denotes the baseline in the different scenarios; FT denotes fine tuning method; BW denotes batch weighting method. Original denotes the original method for SNMT; modified denotes our modified method for UNMT.
		\#1 and \#2 are the results of supervised NMT; others are the results of UNMT. $\mathcal{N}_{in}=10$, $\mathcal{N}_{out} =1$ in original batch weighting method,
		$\mathcal{N}_{in}=1$, $\mathcal{N}_{out} =30$ in modified batch weighting method, and selected pseudo in-domain corpus size is set to 20K for fine tuning method in scenario $IO$ and $IOO$. Note that  $L_2$ in-domain data and all $L_1$ out-of-domain data were used in original fine tuning method for scenario $IOO$.}
	\label{tab:baseline}
\end{table}

Table \ref{tab:baseline} shows the detailed BLEU scores of all UNMT systems on the De$\leftrightarrow$En and Fr$\leftrightarrow$En test sets. \#1 and \#2 are the BLEU scores of SNMT and \#3-to-\#12 are the BLEU scores of UNMT.
Our observations are as follows:

1) The BLEU scores of baselines in the $IIOO$, $IOO$, $IIO$, and $IO$ scenario were presented in the \#5, \#7, \#9, and \#11, respectively. The BLEU scores of UNMT systems after introducing our proposed methods in these scenarios were reported in the \#6, \#8, \#10, and \#12, respectively. 
Compared with original methods, our modified methods are beneficial for improving the performance of UNMT in the defined four scenarios.
%Our proposed methods significantly outperformed baseline systems.% by at most 17 BLEU scores. 
%The proposed methods including fine tuning and batch weighting significantly outperformed the corresponding baseline in every scenario.

2) In the scenario where monolingual training corpora are from same domains, such as $IIOO$, fine tuning method could further improve UNMT performance, achieving an average improvement of 4.8 BLEU scores on all test sets.
%The existing fine tuning method can 

3) In the scenario where monolingual training corpora are from different domains (unique scenario for UNMT domain adaptation), our modified methods achieved average improvements of 4.4, 11.9, and 6.6 BLEU scores in the scenario $IOO$, $IIO$, and $IO$, respectively.%add an average improvement column in Table 3 for easy reading

4)  Our modified batch weighting method improved UNMT performance in the case that there is no out-of-domain monolingual corpora  for  one  language such as scenario $IIO$ and $IO$.
% By contrast， the original batch weighting method achieved worse performance than the baseline. This validates that the supervised domain adaptation method proposed by \newcite{8360031} was not suitable for UNMT. Our proposed batch weighting method could build a more robust UNMT model.
%The details of batch weighting analysis are provided in Appendix~\ref{BW}.
%Data selection  method is  suitable  to  the  case  that  there is no in-domain monolingual corpora  for  one  language  such as scenario $IO$ and $IOO$.
Our modified fine tuning method could further improve translation performance in the case that there is no in-domain monolingual corpora  for  one  language such as scenario $IOO$ and $IO$. %The details of batch weighting and fine tuning analysis are provided in Appendix~\ref{BW} and~\ref{FT}. 
%1) \#1 and \#2 are the BLEU scores of supervised NMT. Our re-implemented supervised NMT system performed better than one reported by \newcite{8360031} in the $II$ scenario.

%2) Others are the BLEU score of UNMT. The BLEU scores in the $II$ and $OO$ scenario were reported in the \#3 and \#4. These two scenarios could be regarded as one of baselines in last four scenarios.

%3) The BLEU scores of baselines in the $IIOO$, $IOO$, $IIO$, and $IO$ scenario were presented in the \#5, \#7, \#9, and \#11, respectively. The BLEU scores of UNMT systems after introducing our proposed methods in these scenarios  were reported in the \#6, \#8, \#10, and \#12, respectively. The proposed methods including fine tuning and batch weighting significantly outperformed  the corresponding baseline in every scenario.
\section{Discussion}
We now further analyze batch weighting and fine tuning methods and perform  an ablation analysis in the unique scenarios for UNMT domain adaptation.
\subsection{Batch Weighting Analysis}
\label{BW}
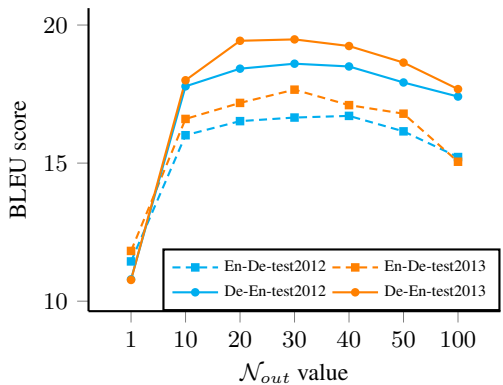
\begin{figure}[ht]
	\setlength{\abovecaptionskip}{0pt}
	\begin{center}
		\scalebox{1}{
			\pgfplotsset{height=5.6cm,width=8.5cm,compat=1.14,every axis/.append style={thick}}
			\begin{tikzpicture}
			\tikzset{every node}=[font=\small]
			\begin{axis}
			[width=7cm,enlargelimits=0.13, tick align=outside, legend style={cells={anchor=west},legend pos=south east, legend columns=2,every axis legend/.append style={
					at={(1,0)}}}, xticklabels={ $1$, $10$,$20$, $30$, $40$,$50$, $100$},
			xtick={0,1,2,3,4,5,6},
			axis y line*=left,
			axis x line*=left,
			ylabel={BLEU score},xlabel={$\mathcal{N}_{out}$ value},font=\small]
			
			\addplot+ [sharp plot,densely dashed,mark=square*,mark size=1.2pt,mark options={solid,mark color=cyan}, color=cyan] coordinates
			{ (0,11.44)(1,16.01)(2,16.52)(3,16.65)(4,16.71)(5,16.15)(6,15.22) };
			\addlegendentry{\tiny En-De-test2012}
			\addplot+ [sharp plot,densely dashed,mark=square*,mark size=1.2pt,mark options={solid,mark color=orange}, color=orange] coordinates
			{ (0,11.82)(1,16.60)(2,17.18)(3,17.66)(4,17.10)(5,16.79)(6,15.05) };
			\addlegendentry{\tiny En-De-test2013}
			
			\addplot+ [sharp plot, mark=*,mark size=1.2pt,mark options={solid,mark color=cyan}, color=cyan] coordinates
			{ (0,10.79)(1,17.78)(2,18.42)(3,18.60) (4,18.50)(5,17.92)(6,17.41)};
			\addlegendentry{\tiny De-En-test2012}
			\addplot+[sharp plot, mark=*,mark size=1.2pt,mark options={solid,mark color=orange}, color=orange] coordinates
			{ (0,10.77)(1,18.00)(2,19.43)(3,19.48) (4,19.24)(5,18.64)(6,17.68)};
			\addlegendentry{\tiny De-En-test2013}
			\end{axis}
			\end{tikzpicture}}
		\caption{\label{fig:lambda}Effect of mini-batch size $\mathcal{N}_{out}$ for UNMT performance after introducing batch weighting method on the En$\leftrightarrow$De dataset in the $IO$ scenario.}
	\end{center}
\end{figure}

In Figure \ref{fig:lambda}, we empirically investigated how the out-of-domain ratio $\mathcal{R}_{out}$ in Eq. (\ref{eq:rout}) affects the UNMT performance on the En$\leftrightarrow$De task in the $IO$ scenario. $\mathcal{N}_{in}$ was set to 1. The selection of $\mathcal{N}_{out}$ influences the  weight of out-of-domain sentences every batch across the entire UNMT training process. Larger values of $\mathcal{N}_{out}$ enable more out-of-domain sentences utilized in the UNMT training. The smaller the value of $\mathcal{N}_{out}$ is, the more important are in-domain sentences. 
As the  Figure \ref{fig:lambda} shows, $\mathcal{N}_{out}$ ranging from 10 to 100 all enhanced UNMT performance and a balanced $\mathcal{N}_{out} = 30$ achieved the best performance.
%In addition, when $\mathcal{N}$ ranged from 30 to 100, all the UNMT performance began to decrease since the much more out-of-domain data .

Moreover, We explored the performance of two batch weighting methods, that is, the existing batch weighting method \cite{8360031} used in NMT domain adaptation and our modified bacth weighting method focused on UNMT domain adaptation.
As shown in Table~\ref{tab:two_batch}, +BW \cite{8360031} ($\mathcal{N}_{in}=10$, $\mathcal{N}_{out} =1$) achieved worse performance than the baseline. Our modified batch weighting method outperformed the baseline by 4.6$\sim$7.2 BLEU scores. This validates that the supervised domain adaptation method proposed by \newcite{8360031} was not suitable for UNMT. Our modified batch weighting method could build a more robust UNMT model.
\begin{table}[ht]
	\centering
	\scalebox{1}{
		\begin{tabular}{lrrrr}
			\toprule
			\multirow{2}{*}{Method} & \multicolumn{2}{c}{De-En} & \multicolumn{2}{c}{En-De} \\
			&test2012 & test2013 &test2012&test2013\\
			\midrule
			Base&10.79 &10.77 &11.44 &11.82 \\
			\midrule
			
			\;\;\;+BW \cite{8360031}&8.15&7.05&9.28&9.70\\
			\;\;\;+BW (our) &17.78  &18.00  & 16.01 & 16.60\\
			\bottomrule
	\end{tabular}}
	\caption{The results of two batch weighting methods in $IO$ scenario on En-De language pairs. }
	\label{tab:two_batch}
\end{table}

\begin{figure}[ht]
	\setlength{\abovecaptionskip}{0pt}
	\begin{center}
		\scalebox{1}{
			\pgfplotsset{height=5.6cm,width=8.5cm,compat=1.14,every axis/.append style={thick},every axis legend/.append style={
					at={(1.7,0.21)}},legend columns=2}
			\begin{tikzpicture}
			\tikzset{every node}=[font=\small]
			\begin{axis}
			[width=7cm,enlargelimits=0.13, tick align=outside, xticklabels={ $5$, $10$,$15$, $20$, $25$,$30$},
			xtick={0,1,2,3,4,5},
			axis y line*=left,
			axis x line*=left,
			ylabel={BLEU score},xlabel={Training time (hour)},font=\small]
			
			\addplot+ [sharp plot,densely dashed,mark=square*,mark size=1.2pt,mark options={solid,mark color=cyan}, color=cyan] coordinates
			{ (0,13.49)(1,14.51)(2,15.07)(3,15.52)(4,15.95)(5,16.01)};
			\addlegendentry{\tiny En-De-BW}
			\addplot+ [sharp plot,densely dashed,mark=square*,mark size=1.2pt,mark options={solid,mark color=brown}, color=brown] coordinates
			{ (0,9.11)(1,10.39)(2,9.01)(3,7.64)(4,7.42)(5,6.75)};
			\addlegendentry{\tiny En-De-Base}
			
			\addplot+ [sharp plot, mark=*,mark size=1.2pt,mark options={solid,mark color=cyan}, color=cyan] coordinates
			{ (0,15.46)(1,16.30)(2,16.49)(3,16.91) (4,17.56)(5,17.78)};
			\addlegendentry{\tiny De-En-BW}
			\addplot+[sharp plot, mark=*,mark size=1.2pt,mark options={solid,mark color=brown}, color=brown] coordinates
			{ (0,8.64)(1,10.07)(2,9.21)(3,8.09) (4,7.98)(5,7.60)};
			\addlegendentry{\tiny De-En-Base}
			\end{axis}
			\end{tikzpicture}}
		\caption{\label{fig:time}The learning curve between baseline and batch weighting model on the En$\leftrightarrow$De test2012 in $IO$ scenario.}
		
	\end{center}
\end{figure}
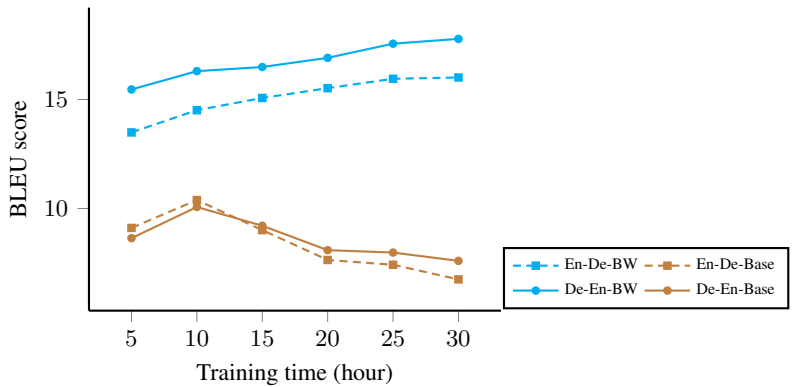

In addition, we also investigated the training time cost between our batch weighting method and the baseline in the $IO$ scenario.
As shown in Figure \ref{fig:time}, both our batch weighting method and the baseline take 30 hours during the whole training process on the $IO$ scenario. The BLEU score of the baseline decreased rapidly after certain epochs due to over-fitting while our proposed batch weight method could continuously improve translation performance during training process. Over the course of training process, our proposed batch weight method performed significantly better than baseline. These demonstrate that our proposed batch weighting method is robust and 
effective. 

\subsection{Fine Tuning Analysis}
\label{FT}
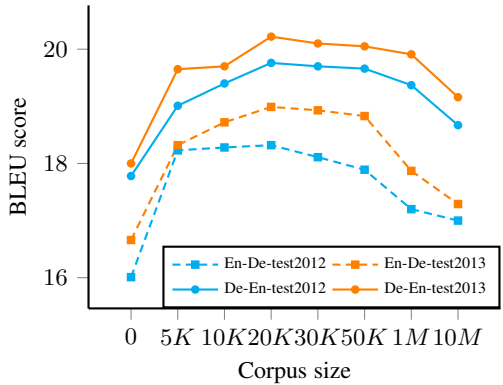
\begin{figure}[ht]
	\setlength{\abovecaptionskip}{0pt}
	\begin{center}
		\scalebox{1}{
			\pgfplotsset{height=5.6cm,width=8.5cm,compat=1.14,every axis/.append style={thick}}
			\begin{tikzpicture}
			\tikzset{every node}=[font=\small]
			\begin{axis}
			[width=7cm,enlargelimits=0.13, tick align=outside, legend style={cells={anchor=west},legend pos=south east, legend columns=2,every axis legend/.append style={
					at={(1,0)}}}, xticklabels={$0$, $5K$,$10K$, $20K$,$30K$,$50K$, $1M$,$10M$},
			xtick={0,1,2,3,4,5,6,7},
			axis y line*=left,
			axis x line*=left,
			ylabel={BLEU score},xlabel={Corpus size},font=\small]
			
			\addplot+ [sharp plot,densely dashed,mark=square*,mark size=1.2pt,mark options={solid,mark color=cyan}, color=cyan] coordinates
			{ (0,16.01)(1,18.23)(2,18.28)(3,18.32)(4,18.11)(5,17.89)(6,17.2)(7,17) };
			\addlegendentry{\tiny En-De-test2012}
			\addplot+ [sharp plot,densely dashed,mark=square*,mark size=1.2pt,mark options={solid,mark color=orange}, color=orange] coordinates
			{ (0,16.66)(1,18.32)(2,18.72)(3,18.99)(4,18.93)(5,18.83) (6,17.87)(7,17.29)};
			\addlegendentry{\tiny En-De-test2013}
			
			\addplot+ [sharp plot, mark=*,mark size=1.2pt,mark options={solid,mark color=cyan}, color=cyan] coordinates
			{ (0,17.78)(1,19.01)(2,19.4)(3,19.76) (4,19.7)(5,19.66)(6,19.37)(7,18.67)};
			\addlegendentry{\tiny De-En-test2012}
			\addplot+[sharp plot, mark=*,mark size=1.2pt,mark options={solid,mark color=orange}, color=orange] coordinates
			{ (0,18)(1,19.65)(2,19.7)(3,20.22) (4,20.1)(5,20.05)(6,19.91)(7,19.16)};
			\addlegendentry{\tiny De-En-test2013}
			\end{axis}
			\end{tikzpicture}}
		\caption{\label{fig:ds}Effect of selected in-domain corpus size for the performance of fine tuning UNMT model on the En$\leftrightarrow$De dataset in the $IO$ scenario.  Corpus size ``0" indicates the result of the UNMT model only with batch weighting method.}
		
	\end{center}
\end{figure}
As shown in Figure \ref{fig:ds}, we empirically investigated how the selected pseudo in-domain corpus size for fine tuning affects the performance of fine tuning UNMT on the En$\leftrightarrow$De task in the $IO$ scenario. 
The larger corpus size brought more pseudo in-domain corpus participate in UNMT further training; the smaller corpus size made pseudo in-domain corpus more precise.
%As Section \ref{method} describes, data selection method works with fine tuning method to improve translation performance.
Corpus size ranging from 5k to 10M all enhanced UNMT performance and UNMT model achieved the best performance when corpus size was set to 20K as shown in Figure \ref{fig:ds}.
This indicates that our modified fine tuning method is robust and effective.

Moreover, we evaluated the different data selection criteria before fine tuning UNMT system on the En$\leftrightarrow$De task  in $IO$ scenario. CED outperformed CE by approximately 1 BLEU score as shown in Table \ref{tab:CE}. This demonstrates that pseudo in-domain corpus selected by CED is more precise for improving UNMT performance.
\begin{table}[ht]
	\centering
	\scalebox{1}{
		\begin{tabular}{lrrrr}
			\toprule
			\multirow{2}{*}{Method} & \multicolumn{2}{c}{De-En} & \multicolumn{2}{c}{En-De} \\
			&test2012 & test2013 &test2012&test2013\\
			\midrule
			CED  &   19.76   &   20.22     &   18.32    &   18.99   \\
			CE  &18.53 & 18.87  &17.19  & 17.81\\
			\bottomrule
	\end{tabular}}
	\caption{Different data selection criteria on the En-De language pairs in $IO$ scenario. CED denotes cross entropy difference criterion $CE_I(s)-CE_O(s)$; CE denotes cross entropy criterion $CE_I(s)$. Pseudo in-domain corpus size is set to 20K.}
	\label{tab:CE}
\end{table}

We also investigated the necessity of denoising auto-encoder during fine tuning process in the $IIOO$ scenario on the En-De language pairs. As shown in Table \ref{tab:FT}, the fine tuning model with denoising performed slightly better than that without denoising. This demonstrates that denoising auto-encoder can further enhance model learning ability during fine-tuning on in-domain data.
\begin{table}[ht]
	\centering
	\scalebox{1}{
		\begin{tabular}{lrrrr}
			\toprule
			\multirow{2}{*}{Method} & \multicolumn{2}{c}{De-En} & \multicolumn{2}{c}{En-De} \\
			&test2012 & test2013 &test2012&test2013\\
			\midrule
			w/o denoising  &   29.80   &   30.99      &   26.39    &   27.84   \\
			w denoising  &29.82 & 31.57  &26.48  & 28.18\\
			\bottomrule
	\end{tabular}}
	\caption{Denoising analysis on the  En-De language pairs in $IIOO$ scenario. }
	\label{tab:FT}
\end{table}

%Table \ref{tab:Ablation3} shows the results in the $IO$ scenario.  +FT  denotes fine tuning was applied to a UNMT baseline system; +BW denotes that  a UNMT system was trained with batch weighting; +BW+FT denotes that fine tuning was applied to a UNMT system trained with batch weighting. 

\subsection{Ablation Analysis}
We performed an ablation analysis to understand the importance of our proposed methods in the $IO$ and $IIO$ scenarios (unique scenarios for UNMT domain adaptation). 

%For the $IIO$ scenario, +BW denotes that  a UNMT system was trained with batch weighting method; +FT denotes that fine tuning was applied to a UNMT baseline system; +FT+BW denotes that fine tuning was applied to a UNMT system trained with batch weighting.
\begin{table}[ht]
	\centering
	\scalebox{1}{
		\begin{tabular}{lrrrr}
			\toprule
			\multirow{2}{*}{Method} & \multicolumn{2}{c}{De-En} & \multicolumn{2}{c}{En-De} \\
			&test2012 & test2013 &test2012&test2013\\
			\midrule
			Base&10.79 &10.77 &11.44 &11.82 \\
			\midrule
			\;\;\;+FT&12.63&12.36&12.22&13.32\\
			\;\;\;+BW& 17.78&18.00 &16.01 &16.60\\
			\;\;\;+FT+BW  & 19.76 &  20.22 & 18.32 &18.99\\
			\bottomrule
	\end{tabular}}
	\caption{Ablation analysis on the En$\leftrightarrow$De dataset in $IO$ scenario. +BW denotes that  a UNMT system was trained with batch weighting method; +FT denotes that fine tuning was applied to a UNMT baseline system; +FT+BW denotes that fine tuning was applied to a UNMT system trained with batch weighting. }
	\label{tab:Ablation3}
\end{table}
As shown in Table \ref{tab:Ablation3}, we observed that both of +FT and +BW outperformed the Base in the $IO$ scenario and +BW was more suitable for this scenario, achieving much more improvement in BLEU score. Moreover, +FT+BW can complement each other to further improve UNMT performance, achieving the best performance in the $IO$ scenario.

\begin{table}[ht]
	\centering
	\scalebox{1}{
		\begin{tabular}{lrrrr}
			\toprule
			\multirow{2}{*}{Method} & \multicolumn{2}{c}{De-En} & \multicolumn{2}{c}{En-De} \\
			&test2012 & test2013 &test2012&test2013\\
			\midrule
			Base&11.11 &10.30 &11.54 &11.95 \\
			\midrule
			\;\;\;+BW &18.96  &18.87  & 20.23 & 20.81\\
			\;\;\;+FT& 19.78&20.70 &17.24 &18.02\\
			\;\;\;+FT+BW  &26.12  & 27.33  &22.63  & 23.72\\
			\bottomrule
	\end{tabular}}
	\caption{Ablation analysis on the En$\leftrightarrow$De dataset in $IIO$ scenario.}
	\label{tab:Ablation}
\end{table}
As shown in Table~\ref{tab:Ablation},  we observed that both of +FT and +BW outperformed the Base in the $IIO$ scenario. 
In particular, the +FT+BW was further better than both +FT and +BW.  
This means that our modified batch weighting and fine tuning methods can improve the performance of UNMT in this $IIO$ scenario, especially, both of them can complement each other to further improve translation performance.
\section{Related Work}

Recently, UNMT \cite{DBLP:journals/corr/abs-1710-11041,lample2017unsupervised,P18-1005}, that has been trained via bilingual word embedding initialization, denoising auto-encoder, and back-translation and sharing latent representation mechanisms, has attracted great interest in the machine translation community.
\newcite{lample2018phrase}  achieved remarkable results on some similar language pairs by concatenating  two bilingual corpora as one monolingual corpus and using monolingual embedding to initialize the embedding layer of UNMT. 
\newcite{wu-etal-2019-extract} proposed an extract-edit approach,  to extract and then edit real sentences from the target monolingual corpora instead of back-translation. 
\newcite{sun-etal-2019-unsupervised} proposed bilingual word embedding agreement mechanisms to improve UNMT performance. %, achieving better UNMT performance.
More recently, \newcite{DBLP:journals/corr/abs-1901-07291} achieved state-of-the-art UNMT performance by introducing the pretrained cross-lingual language model. However, previous work only focuses  on  how  to  build  state-of-the-art  UNMT systems on specific domain and ignore the effect of UNMT on different domain.
Research on domain adaptation for UNMT  has  been  limited   while domain  adaptation  methods  have  been  well-studied  in SNMT. 

\newcite{DBLP:conf/coling/ChuW18} gave a survey of domain adaptation techniques for SNMT. Domain adaptation for SNMT could be categorized into two main categories: data optimization and model optimization. Data optimization methods included  synthetic parallel corpora generation using in-domain monolingual corpus \cite{P16-1009,hu-etal-2019-domain} and data selection for out-of-domain parallel corpora \cite{wang-etal-2017-sentence,DBLP:conf/emnlp/WeesBM17,zhang-etal-2019-curriculum}. Training objective optimization including instance weighting \cite{wang-etal-2017-instance,DBLP:conf/aclnmt/ChenCFL17} and fine tuning~\cite{Luong-Manning:iwslt15,P16-1009,DBLP:journals/corr/FreitagA16,DBLP:journals/corr/ServanCS16,chu-etal-2017-empirical}, architecture optimization~\cite{DBLP:conf/ranlp/KobusCS17,britz-etal-2017-effective,gu-etal-2019-improving} and decoding optimization \cite{DBLP:journals/corr/FreitagA16,khayrallah-etal-2017-neural,saunders-etal-2019-domain} were common  model optimization methods for domain adaptation. 
\section{Conclusion}
In this paper, we mainly raise the issue of UNMT domain  adaptation since domain adaptation methods for UNMT have never been proposed. We empirically show different scenarios for domain-specific UNMT. %Some scenarios are unique scenarios for UNMT domain adaptation. 
Based on these scenarios, we revisit the effect of the existing domain adaptation methods including batch weighting and fine tuning methods in UNMT. Experimental results show our modified  corresponding methods  improve the performance of UNMT in these scenarios. In the future, we will try to investigate other unsupervised domain adaptation methods to further improve domain-specific UNMT performance.

\bibliographystyle{coling}
\bibliography{coling2020}

\end{document}